# Classical Statistics and Statistical Learning in Imaging Neuroscience

Danilo Bzdok[1,2,3]

1 Department of Psychiatry, Psychotherapy and Psychosomatics, Medical Faculty, RWTH Aachen, Germany
2 JARA, Translational Brain Medicine, Aachen, Germany
3 Parietal team, INRIA, Neurospin, bat 145, CEA Saclay, 91191 Gif-sur-Yvette, France

Prof. Dr. Dr. Danilo Bzdok
Department for Psychiatry, Psychotherapy and Psychosomatics
Pauwelsstraße 30
52074 Aachen
Germany
mail: danilo[DOT]bzdok[AT]rwth-aachen[DOT]de

**Short Abstract:** Neuroimaging research has predominantly drawn conclusions based on classical statistics. Recently, statistical learning methods enjoy increasing popularity. These methodological families used for neuroimaging data analysis can be viewed as two extremes of a continuum, but are based on different histories, theories, assumptions, and outcome metrics; thus permitting different conclusions. This paper portrays commonalities and differences between classical statistics and statistical learning with regard to neuroimaging research. The conceptual implications are illustrated in three case studies. It is thus tried to resolve possible confusion between classical hypothesis testing and data-guided model estimation by discussing ramifications for the neuroimaging access to neurobiology.

**Long Abstract:** Neuroimaging research has predominantly drawn conclusions based on classical statistics, including null-hypothesis testing, *t*-tests, and ANOVA. Throughout recent years, statistical learning methods enjoy increasing popularity, including cross-validation, pattern classification, and sparsity-inducing regression. These two methodological families used for neuroimaging data analysis can be viewed as two extremes of a continuum. Yet, they originated from different historical contexts, build on different theories, rest on different assumptions, evaluate different outcome metrics, and permit different conclusions. This paper portrays commonalities and differences between classical statistics and statistical learning with their relation to neuroimaging research. The conceptual implications are illustrated in three common analysis scenarios. It is thus tried to resolve possible confusion between classical hypothesis testing and data-guided model estimation by discussing their ramifications for the neuroimaging access to neurobiology.

**Main Text**

"The trick to being a scientist is to be open to using a wide variety of tools."

Leo Breiman (2001)

**1. Introduction**

Among the greatest challenges humans face are cultural misunderstandings between individuals, groups, and institutions (Hall, 1989). The topic of the present paper is the culture clash between statistical inference by null-hypothesis rejection and out-of-sample generalization (L. Breiman, 2001; Friedman, 1998; Shmueli, 2010) that are increasingly combined in the brain-imaging domain (N. Kriegeskorte, Simmons, Bellgowan, & Baker, 2009). Ensuing inter-cultural misunderstandings are unfortunate, because the invention and application of new research methods has always been a driving force in the neurosciences (Deisseroth, 2015; Greenwald, 2012; Yuste, 2015). It is the goal of the present paper to disentangle classical inference and generalization inference by juxtaposing their historical trajectories (section 2), modelling philosophies (section 3), conceptual frameworks (section 4), and performance metrics (section 5).

During the past 15 years, neuroscientists have transitioned from exclusively qualitative reports of few patients with neurological brain lesions to quantitative lesion-symptom mapping on the voxel level in hundreds of patients (Bates et al., 2003). We have gone from manually staining and microscopically inspecting single brain slices to 3D models of neuroanatomy at micrometer scale (Amunts et al., 2013). We have also gone from individual experimental studies to the increasing possibility of automated knowledge aggregation across thousands of previously isolated neuroimaging findings (T. Yarkoni, Poldrack, Nichols, Van Essen, & Wager, 2011). Rather than laboriously collecting and publishing in-house data in a single paper, investigators are now routinely reanalyzing multi-modal data repositories managed by national, continental, and inter-continental consortia (Kandel, Markram, Matthews, Yuste, & Koch, 2013; Henry Markram, 2012; Poldrack & Gorgolewski, 2014; Van Essen et al., 2012). The granularity of neuroimaging datasets is hence growing in terms of scanning resolution, sample size, and complexity of meta-information (S. Eickhoff, Turner, Nichols, & Van Horn, 2016; Van Horn & Toga, 2014). As an important consequence, the scope of neuroimaging analyses has expanded from the predominance of null-hypothesis testing to statistical-learning methods that are i) more data-driven by flexible models, ii) naturally scalable to high-dimensional data, and iii) more heuristic by increased reliance on numerical optimization (Jordan & Mitchell, 2015; LeCun, Bengio, & Hinton, 2015). *Statistical learning* (T. Hastie, Tibshirani, & Friedman, 2001) henceforth comprises the umbrella of "machine learning", "data mining", "pattern recognition", "knowledge discovery", and "high-dimensional statistics".

In fact, the very notion of *statistical inference* may be subject to expansion. The following definition is drawn from a committee report to the National Academies of the USA (Jordan et al., 2013, p. 8): "Inference is the problem of turning data into knowledge, where knowledge often is expressed in terms of variables [...] that are not present in the data per se, but are present in models that one uses to interpret the data." According to this authoritative definition, statistical inference can be understood as encompassing not only classical null-hypothesis falsification but also out-of-sample generalization (cf. Jacob Cohen, 1990; G. Gigerenzer & Murray, 1987). Classical statistics and statistical learning might give rise to different categories of inference, which remains an inherently difficult concept to define (Chamberlin, 1890; Pearl, 2009; Platt, 1964). Any choice of statistical method for a neurobiological investigation predetermines the spectrum of possible results and permissible conclusions.

Briefly taking an epistemological perspective, a new *scientific fact* is probably not established in vacuo (Fleck, Schäfer, & Schnelle, 1935; italic terms in this passage taken from source). Rather, the *object* is recognized and accepted by the *subject* according to socially conditioned *thought styles* cultivated among members of *thought collectives*. A witnessed and measured neurobiological phenomenon tends to only become "true" if not at odds with the constructed *thought history* and *closed opinion system* shared by that subject. In the following, two such *thought milieus* will be revisited and reintegrated in the context of imaging neuroscience: classical statistics (CS) and statistical learning (SL).

## 2. Different histories

### 2.1 The origins of classical hypothesis testing and learning algorithms

The largely independent historical trajectories of the two statistical families are even evident from their most basic terminology. Inputs to statistical models are usually called *independent variables* or *predictors* in the CS literature, but are called *features* collected in a *feature space* in the SL literature. The outputs are called *dependent variables* or *responses* in CS and *target variables* in SL, respectively.

Around 1900 the notions of standard deviation, goodness of fit, and the "p < 0.05"-threshold emerged (Cowles & Davis, 1982). This was also the period when William S. Gosset published the *t*-test under the incognito name "Student" to quantify production quality in Guinness breweries. Motivated by concrete problems such as the interaction between potato varieties and fertilizers, Ronald A. Fisher invented the analysis of variance (ANOVA), null-hypothesis testing, promoted p values, and devised principles of proper experimental conduct (R. A. Fisher, 1925; Ronald A. Fisher, 1935; Ronald A. Fisher & Mackenzie, 1923). An alternative framework for hypothesis testing was proposed by Jerzy Neyman and Egon S. Pearson, which introduced the statistical notions of power, false positives and false negatives, but left out the concept of p values (Neyman & Pearson, 1933). This was a time when controlled experiments were preferentially performed on single individuals, before the gradual transition to participant groups in the 30s and 40s, and before electrical calculators emerged after World War II (Efron & Tibshirani, 1991; Gerd Gigerenzer, 1993). Student's *t*-test and Fisher's inference framework were institutionalized by American psychology textbooks that were widely read in the 40s and 50s. Neyman and Pearson's approach only became increasingly known in the 50s and 60s. This led authors of social science textbooks to promote a somewhat incoherent mixture of the Fisher and Neyman-Pearson approaches to statistical inference, typically without explicit mention. It is this conglomerate of classical frameworks to performing inference that today's textbooks of applied statistics have inherited (Bortz, 2006; Moore & McCabe, 1989).

It is a topic of current debate[1,2,3] whether CS is a discipline that is separate from SL (e.g., L. Breiman, 2001; Chambers, 1993; Friedman, 2001) or if statistics is a broader class that includes CS and SL as its members (e.g., Cleveland, 2001; Jordan & Mitchell, 2015; Tukey, 1962). SL methods are frequently adopted by computer scientists, physicists, engineers, and others who have no formal statistical background and are typically working in industry rather than academia. In fact, John W. Tukey foresaw many of the developments that led up to what one might today call statistical learning (Tukey, 1962, 1965). He proposed a "peaceful collision of computing and statistics" as well as the distinction into "exploratory" and "confirmatory" data analysis[4]. This emphasized data-driven analysis techniques as a toolbox useful in a large variety of real-world settings to gain an intuition of the data properties. Kernel methods, neural networks, decision trees, nearest neighbors, and graphical models all actually originated in the CS community, but mostly continued to develop in the SL community (Friedman, 2001). As often cited beginnings of self-learning algorithms, the *perceptron* was an early brain-inspired computing algorithm (Rosenblatt, 1958), and Arthur Samuel created a checker board program that succeeded in beating its own creator (Samuel, 1959). Such studies on *artificial intelligence* (AI) led to enthusiastic optimism and subsequent disappointment due to the slow progress of learning algorithms. The consequence was a slow-down of research, funding, and interest during the so-called "AI winters" in the late 70s and around the 90s (D. D. Cox & Dean, 2014; Kurzweil, 2005; Russell & Norvig, 2002), while the increasingly available computers in the 80s encouraged a new wave of statistical algorithms (Efron & Tibshirani, 1991). The difficult-to-train but back then widely used neural network algorithms were superseded by support vector machines with convincing out-of-the-box performances (Cortes & Vapnik, 1995). Later, the use of SL methods increased steadily in many quantitative scientific domains as they underwent an increase in information granularity from classical "long data" (samples n > variables p) to modern "wide data" (n < p) (Tibshirani, 1996). The emerging field of SL has been much conceptually consolidated by the seminal book "The Elements of Statistical Learning" (T. Hastie et al., 2001). The coincidence of changing data properties, increasing computational power, and cheaper memory resources encouraged a resurge in SL research and applications approximately since 2000 (House of Common, 2016; Manyika et al., 2011). Over the last 15 years, *sparsity* assumptions gained increasing relevance for statistical tractability and domain interpretability when using *supervised* and *unsupervised* learning algorithms (i.e., with and without target variables) by imposing a prior distribution on the model parameters (Bach, Jenatton, Mairal, & Obozinski, 2012). According to the "bet on sparsity" (Trevor Hastie, Tibshirani, & Wainwright, 2015), only a subset of the features should be expected to be relevant because no existing statistical method performs well in the *dense* high-dimensional scenario that assumes all features to be relevant in the "true" model (Brodersen, Haiss, et al., 2011; T. Hastie et al., 2001). This enabled reproducible and interpretable statistical relationships in the high-dimensional "n << p" regime (Bühlmann & Van De Geer, 2011; Trevor Hastie et al., 2015). More recently, improvements in training very "deep" (i.e., many non-linear hidden layers) neural networks architectures (Geoffrey E. Hinton & Salakhutdinov, 2006) have much improved automatized feature selection (Bengio, Courville, & Vincent, 2013) and have exceeded human-level performance in several tasks (LeCun et al., 2015). For instance, one recent deep reinforcement learning algorithm mastered playing 49 different computer games based on simple pixel input alone (Mnih et al., 2015). Today, systematic education in SL is still rare at most universities, in contrast to the omnipresence of CS courses (Burnham & Anderson, 2014; Cleveland, 2001; Donoho, 2015; Vanderplas, 2013).

---

[1] "Data Science and Statistics: different worlds?" (Panel at Royal Statistical Society UK, March 2015) (https://www.youtube.com/watch?v=C1zMUjHOLr4)

[2] "50 years of Data Science" (David Donoho, Tukey Centennial workshop, USA, Sept. 2015)

[3] "Are ML and Statistics Complementary?" (Max Welling, 6th IMS-ISBA meeting, December 2015)

[4] As a very recent reformulation of the same idea: "If the inference/algorithm race is a tortoise-and-hare affair, then modern electronic computation has bred a bionic hare." (Efron & Hastie, 2016)

**2.2 Related spotlights in the history of neuroimaging analysis methods**

For more than a century, neuroscientific conclusions were mainly drawn from brain lesion reports (Broca, 1865; Harlow, 1848; Wernicke, 1881), microscopical inspection (Brodmann, 1909; Vogt & Vogt, 1919), brain stimulation during surgery (Penfield & Perot, 1963), and pharmacological intervention (Clark, Del Giudice, & Aghajanian, 1970), often without strong reliance on statistical methodology. The advent of more readily quantifiable neuroimaging methods (Fox & Raichle, 1986) then allowed for in-vivo characterization of the neural correlates underlying sensory, cognitive, or affective tasks. Ever since, topographical localization of neural activity increases and decreases was dominated by analysis approaches from CS, especially the general linear model (GLM; dates back to Nelder & Wedderburn, 1972). Although the GLM is well known not to correspond to neurobiological reality, it has provided good and interpretable approximations. It was and still is routinely used in a mass-univariate regime that computes simultaneous univariate statistics for each independent voxel observation of brain scans (Friston et al., 1994). It involves fitting beta coefficients corresponding to the columns of a *design matrix* (i.e., prespecified stimulus/task/behavior indicators, the independent variables) to a single voxel's imaging time series of measured neural activity changes (i.e., dependent variable) to obtain a beta coefficient for each indicator. It is seldom mentioned that the GLM would not have been solvable for unique solutions in the high-dimensional regime because the number of input variables p exceeded by far the number of samples n (i.e., under-determined system of equations), which incapacitates many statistical estimators from CS (cf. Giraud, 2014; Trevor Hastie et al., 2015). Regularization by sparsity-inducing norms, such as in modern regression analysis via Lasso and ElasticNet (cf. Trevor Hastie et al., 2015; Jenatton, Audibert, & Bach, 2011), emerged only later as a principled way to de-escalate the need for dimensionality reduction and to enable the tractability of the high-dimensional "p > n" case (Tibshirani, 1996). Many software packages for neuroimaging analysis consequently implemented discrete voxel-wise analyses with classical inference. The ensuing multiple comparisons problem motivated more than two decades of methodological research (Friston, 2006; Thomas E. Nichols, 2012; Stephen M. Smith, Matthews, & Jezzard, 2001; Worsley, Evans, Marrett, & Neelin, 1992). It was initially addressed by reporting uncorrected (Vul, Harris, Winkielman, & Pashler, 2008) or Bonferroni-corrected findings (Thomas E.

Nichols, 2012), then increasingly by false discovery rate (Genovese, Lazar, & Nichols, 2002) and cluster-level tresholding (Stephen M. Smith & Nichols, 2009). Further, it was early acknowledged that the unit of interest should be spatially neighboring voxel groups and that the null hypothesis needed to account for all voxels exhibiting some signal (Chumbley & Friston, 2009). These concerns were addressed by inference in locally smooth neighborhoods based on *random field theory* that models discrete voxel activations as topological units with continuous activation height and extent (Worsley et al., 1992). That is, the spatial dependencies of voxel observations were not incorporated into the GLM estimation step, but instead during the subsequent model inference step to alleviate the multiple comparisons problem.

To abandon the voxel-independence assumption of the mass-univariate GLM, SL models were proposed early on for neuroimaging investigations. For instance, principal component analysis was used to distinguish local and global neural activity changes (Moeller, Strother, Sidtis, & Rottenberg, 1987) as well as to study Alzheimer's disease (Grady et al., 1990), while canonical correlation analysis yielded complex relationships between task-free neural activity and schizophrenia symptoms (Friston, Liddle, Frith, Hirsch, & Frackowiak, 1992). Note that these first approaches to "multivariate" brain-behavior associations did not ignite a major research trend (cf. Friston et al., 2008; Worsley, Poline, Friston, & Evans, 1997). Supervised classification estimators were also used in early structural neuroimaging analyses (Herndon, Lancaster, Toga, & Fox, 1996) and they improved preprocessing performance for volumetric neuroimaging data (Ashburner & Friston, 2005). However, the popularity of SL methods only peaked after being rebranded as "mind-reading", "brain decoding", and "multivariate pattern analysis" that appealed by identifying ongoing thought from neural activity alone (J. D. Haynes & Rees, 2005; Kamitani & Tong, 2005). Up to that point, the term *prediction* had less often been used in the concurrent sense of out-of-sample generalization of a learning model and more often in the incompatible sense of in-sample linear correlation between (time-free or time-shifted) data (Gabrieli, Ghosh, & Whitfield-Gabrieli, 2015; Shmueli, 2010). The "searchlight" approach for "pattern-information analysis" subsequently enabled *whole-brain* assessment of *local* neighborhoods of predictive patterns of neural activity fluctuations (N. Kriegeskorte, Goebel, & Bandettini, 2006). The position of such "decoding" models within the branches of statistics has seldom been made explicit (but see Friston et al., 2008). The growing interest was manifested in first review and tutorial papers that were published on applying SL methods to neuroimaging data (J. D. Haynes & Rees, 2006; Mur, Bandettini, & Kriegeskorte, 2009; Pereira, Mitchell, & Botvinick, 2009). The conceptual appeal of this new access to the neural correlates of cognition and behavior was flanked by availability of the necessary computing power and memory resources. This was also a precondition for regularization by structured sparsity penalties (Wainwright, 2014) that incorporate neurobiological priors as local spatial dependence (Gramfort, Thirion, & Varoquaux, 2013; Michel, Gramfort, Varoquaux, Eger, & Thirion, 2011) or spatial-temporal dependence (Gramfort, Strohmeier, Haueisen, Hamalainen, & Kowalski, 2011). Although challenging, "deep" neural networks have recently been introduced to neuroimaging (de Brebisson & Montana, 2015; Güçlü & van Gerven, 2015; Plis et al., 2014). The application of these "deep" statistical architectures occurs in an atheoretical, more empirically justified setting as their mathematical properties are incompletely understood and many formal guarantees are missing (but see Bach, 2014). Nevertheless, they might help in deciphering and approximating the nature of neural processing in the brain (D. D. Cox & Dean, 2014; Yamins & DiCarlo, 2016). This agenda appears closely related to David Marr's distinction of information processing into computational (~what?), algorithmic-representational (~how?), and implementational-physical (~where?) levels (Marr, 1982). Last but not least, there is an always bigger interest in and pressure for data sharing, open access, and building "big-data" repositories in neuroscience (Devor et al., 2013; Gorgolewski et al., 2014; Kandel et al., 2013; Henry Markram, 2012; Poldrack & Gorgolewski, 2014; Van Essen et al., 2012). As the dimensionality and complexity of neuroimaging datasets increases, neuroscientific investigations will probably benefit increasingly from SL methods and their variants adapted to the data-intense regime (e.g., Engemann & Gramfort, 2015; Kleiner, Talwalkar, Sarkar, & Jordan, 2012; Zou, Hastie, & Tibshirani, 2006). While larger data quantities allow detection of more subtle effects, false positive findings are likely to become a major issue as they unavoidably arise from statistical estimation in the high-dimensional data scenario (Jordan et al., 2013; Meehl, 1967).

## 3. Different philosophies

### 3.1 Two different modelling goals in statistical methods

One of the possible ways to catalogue statistical methods is by framing them along the lines of classical statistics and statistical learning. Statistical methods can thus be conceptualized as spanning a continuum between the two poles of CS and SL (Jordan et al., 2013; p. 61). While some statistical methods cannot easily be categorized by this distinction, the two families of statistical methods can generally be distinguished by a number of representative properties.

As the precise relationship between CS and SL has seldom been explicitly characterized in mathematical terms, the author must resort to more descriptive explanations (see already Efron, 1978). One of the key differences becomes apparent when thinking of the neurobiological phenomenon under study as a black box (L. Breiman, 2001). CS typically aims at modelling the black box by making a set of accurate assumptions about its content, such as the nature of the signal distribution. Gaussian distributional assumptions have been very useful in many instances to enhance mathematical convenience and, hence, computational tractability. SL typically aims at finding any way to model the output of the black box from its input while making the least assumptions possible (Abu-Mostafa, Magdon-Ismail, & Lin, 2012). In CS the stochastic processes that generated the data is therefore treated as partly known, whereas in SL the phenomenon is treated as complex, largely unknown, and partly unknowable. In this sense, CS tends to be more *analytical* by imposing mathematical rigor on the phenomenon, whereas SL tends to be more *heuristic* by finding useful approximations to the phenomenon. CS specifies the properties of a given statistical model at the beginning of the investigation, whereas in SL there is a bigger emphasis on models whose parameters and at times even structures (e.g., learning algorithms creating decision trees) are generated during the statistical estimation. The SL-minded investigator may favor simple, tractable models even when making conscious false assumptions (Domingos, 2012) because sufficiently large data quantities are expected to remedy them (Devroye, Györfi, & Lugosi, 1996; Halevy, Norvig, &

Pereira, 2009). A new function with potentially thousands of parameters is created that can predict the output from the input alone, without explicit programming model. This requires the input features to represent different variants of all relevant configurations of the examined phenomenon in nature. In CS the mathematical assumptions are typically stated explicitly, while SL models frequently have implicit assumptions that may be less openly discussed. Intuitively, the truth is believed to be in the model (cf. Wigner, 1960) in a CS-constrained statistical regime, while it is believed to be in the data (cf. Halevy et al., 2009) in a SL-constrained statistical regime. Also differing in their output, CS typically yields *point and interval estimates* (e.g., p values, variances, confidence intervals), whereas SL frequently outputs *functions* (e.g., the k-means centroids or a trained classifier's decision function can be applied to new data). As another tendency, SL revolves around solving and introducing prior knowledge into iterative numerical optimization problems. CS methods are more often closed-form one-shot computations without any successive approximation process, although several models are also fitted numerically by maximum likelihood estimation (Boyd & Vandenberghe, 2004; Jordan et al., 2013).

In more formal terms, CS relates closely to statistics for *confirmatory data analysis*, whereas SL relates more to statistics for *exploratory data analysis* (Tukey, 1962). In practice, CS is probably more often applied to *experimental data,* where a set of target variables are systematically controlled by the investigator and the system under studied has been subject to structured perturbation. Instead, SL is perhaps more typically applied to *observational data* without such structured influence and the studied system has been left unperturbed (Domingos, 2012). From yet another important angle, one should not conflate so-called *explanatory modelling* and *predictive modelling* (Shmueli, 2010). CS mainly performs retrospective *explanatory modelling* by emphasis on the operationalization of preselected hypothetical constructs as measurable outcomes using frequently linear, interpretable models. SL mainly performs prospective *predictive modelling* and quantifies the generalization to future observations with or without formal incorporation of hypothetical concepts into models that are more frequently non-linear with challenging to impossible interpretability. There is the *often-* *overlooked misconception that models with high explanatory power do necessarily exhibit high predictive power* (Lo, Chernoff, Zheng, & Lo, 2015; Wu, Chen, Hastie, Sobel, & Lange, 2009). An important outcome measure in CS is the quantified *significance* associated with a statistical relationship between few variables given a pre-specified model. The outcome measure for SL is the quantified *generalizability* or *robustness of patterns* between many variables or, more generally, the robustness of special structure in the data (T. Hastie et al., 2001). CS tends to *test for a particular structure* in the data based on *analytical guarantees*, such as mathematical convergence theorems about approximating the population properties with increasing sample size. Instead, SL tends to *explore particular structure* in the data and quantify its *generalization* to new data certified by *empirical guarantees*, such as by explicit evaluation of the predictiveness of a fitted model to unseen data (Efron & Tibshirani, 1991). CS thus resorts more to what Leo Breiman called *data modelling*, imposing an a-priori model in a top-down fashion, while SL would be *algorithmic modelling*, fitting a model as a function of the data at hand in a bottom-up fashion (L. Breiman, 2001).

Although this polarization of statistical methodology may be oversimplified, it serves as a didactic tool to confront two different perspectives. Taken together, CS was mostly fashioned for problems with small samples that can be grasped by plausible models with a small number of parameters chosen by the investigator in an analytical fashion. SL was mostly fashioned for problems with many variables in potentially large samples with rare knowledge of the data-generating process that are emulated by a mathematical function created from data by a machine in a heuristic fashion. Tests from CS therefore typically assume that the data behave according to known mechanisms, whereas SL exploits algorithmic techniques to avoid a-priori specifications of data mechanism. In this way, CS preassumes and tests a model *for the data*, whereas SL learns a model *from the data* [5].

---

[5] "Indeed, the trend since Pearl's work in the 1980's has been to blend reasoning and learning: put simply, one does not need to learn (from data) what one can infer (from the current model). Moreover, one does not need to infer what one can learn (intractable inferential procedures can be circumvented by collecting data)." (Jordan, 2010)

Obviously, the existing repertoire of statistical methods could have been dissected in different ways. For instance, the *Bayesian-frequentist* distinction is orthogonal to the CS-SL distinction (Efron, 2005; Freedman, 1995; Z. Ghahramani, 2015). Bayesian statistics can be viewed as subjective and optimistic in considering the data at hand conditioned on preselected distributional assumptions for the probabilistic model to perform direct inference (Wagenmakers, Lee, Lodewyckx, & Iverson, 2008). Frequentist statistics are more objective and pessimistic in considering a distribution of possible data with unknown model parameters to perform indirect inference by the differences between observed and model-derived data. It is important to appreciate that Bayesian statistics can be adopted in both CS and SL families in various flavors (Friston et al., 2008; Geoffrey E. Hinton, Osindero, & Teh, 2006; Kingma & Welling, 2013). Furthermore, CS and SL approaches can hardly be clearly categorized as either *discriminative* or *generative* (Bishop, 2006; G.E. Hinton, Dayan, Frey, & Neal, 1995; Jordan, 1995; Ng & Jordan, 2002). Discriminative models focus on solving a supervised problem of predicting a class y by directly estimating $P(y|X)$ without capturing special structure in the data X. Typically more demanding in data quantity and computational resources, generative models estimate special structure by $P(y|X)$ from $P(X|y)$ and $P(y)$ and can thus produce synthetical examples $\tilde{X}$ for each class y. Further, CS and SL cannot be disambiguated into *deterministic* versus *probabilistic* models (Shafer, 1992). Statistical models are often not exclusively deterministic because they incorporate a component that accounts for unexpected, noisy variation in the data. Each probabilistic model can also be viewed as a superclass of a deterministic model (Norvig, 2011). Neither can the terms *univariate* and *multivariate* be exclusively grouped into either CS or SL. They traditionally denote reliance on *one* versus *several* dependent variables (CS) or target variables (SL) in the statistical literatures. In the neuroimaging literature, "multivariate" however frequently refers to high-dimensional approaches operating on the neural activity from all voxels in opposition to mass-univariate approaches operating on single voxel activity (Brodersen, Haiss, et al., 2011; Friston et al., 2008). CS can be divided into uni- and multivariate groups of statistical tests (Ashburner & Kloppel, 2011; Friston et al., 2008), while SL is largely focused on higher-dimensional problems that often naturally transform to the multivariate classification/regression setting. Tapping on yet another terminology, CS methods may be argued to be closer to *parametric statistics* by instantiating statistical models whose number of parameters is fixed, finite, and not a function of sample size (Bishop, 2006; Z. Ghahramani, 2015). Instead, SL methods are more often (but not exclusively) realized in a *non-parametric setting* by instantiating flexible models whose parameters grow increase explicitly or implicitly with more input data. Parametric approaches are often more successful if few observations are available. Conversely, non-parametric approaches may be more naturally prepared to capture emergence properties that only occur from larger datasets (Z. Ghahramani, 2015; Halevy et al., 2009; Jordan et al., 2013). For instance, a non-parametric (but not parametric) classifier can extract ever more complex decision boundaries from increasing training data (e.g., decision tress and nearest neighbors). Most importantly, neither CS nor SL can generally be considered superior. This is captured by the *no free lunch theorem*[6] (Wolpert, 1996), which states that no single statistical strategy can consistently do better in all circumstances (cf. Gerd Gigerenzer, 2004). The investigator has the discretion to choose which statistical approach best suited to the neurobiological phenomenon under study and the neuroscientific research object at hand.

**3.2 Different modelling goals in neuroimaging**

Statistical analysis grounded in CS and SL is more closely related to *encoding models* and *decoding models* in the neuroimaging domain, respectively (Nikolaus Kriegeskorte, 2011; T. Naselaris, Kay, Nishimoto, & Gallant, 2011; Pedregosa, Eickenberg, Ciuciu, Thirion, & Gramfort, 2015; but see Güçlü et al, 2015). Encoding models regress the data against a design matrix with potentially many explanatory columns of stimulus (e.g., face versus house pictures), task (e.g., to evalute or to attend), or behavioral (e.g., age or gender) indicators by fitting general linear models. In contrast, decoding models typically predict these indicators by training and testing classification algorithms on different splits from the whole dataset. In CS parlance, the encoding model fits the neural activity data by the *beta coefficients*, the *dependent variables*, according to the indicators in the *design matrix* columns, the *independent variables*. An explanation for decoding models in SL jargon would be that the *model weights* of a *classifier* are fitted on the *training set* of the *input data* to *predict* the *class labels*, the *target variables,* and are subsequently evaluated on the *test set* by *cross-validation* to obtain their *out-of-sample generalization performance.* Put differently, a GLM fits coefficients of stimulus/task/behavior indicators on neural activity data for each voxel separately given the design matrix (T. Naselaris et al., 2011). Classifiers predict entries of the design matrix for all voxels simultaneously given the neural activity data (Pereira et al., 2009). A key difference between CS-mediated encoding models and SL-mediated decoding models thus pertains to the *direction of inference* between brain space and indicator space (Friston et al., 2008; Varoquaux & Thirion, 2014). The mapping direction hence pertains to the question whether the indicators in the model act as *causes* by representing deterministic experimental variables of an encoding model or *consequences* by representing probabilistic outputs of a decoding model (Friston et al., 2008). These considerations also reveal the intimate relationship of CS models to the notion of so-called *forward inference,* while SL methods are probably more often used for formal *reverse inference* in functional neuroimaging (S. B. Eickhoff et al., 2011; Poldrack, 2006; Varoquaux & Thirion, 2014; T. Yarkoni et al., 2011). On the one hand, *forward inference* relates to encoding models by testing the probability of observing activity in a brain location given knowledge of a psychological process. *Reverse inference*, on the other hand, relates to brain "decoding" by testing the probability of a psychological process being present given knowledge of activation in a brain location. Finally, establishing a brain-behavior association has been argued to be more important than the actual direction of the mapping function. This is because "showing that one can decode activity in the visual cortex to classify [...] a subject's percept is exactly the same as demonstrating significant visual cortex responses to perceptual changes" and, conversely, "all demonstrations of functionally specialized responses represent an implicit mindreading" (Friston, 2009).

---

[6] In the supervised setting, there is no a priori distinction between learning algorithms evaluated by out-of-sample prediction error. In the optimization setting of finite spaces, all algorithms searching an extremum perform identical when averaged across possible cost functions. (http://www.no-free-lunch.org/)

More specifically, GLM encoding models follow a *representational agenda* by testing hypotheses on *regional effects of functional specialization* in the brain (where?). A *t*-test is used to compare pairs of measures to statistically distinguish one target and one non-target set of mental operations (Friston et al., 1996). Essentially, this formal tests for significant differences between the beta coefficients corresponding to two stimulus/task indicators based on well-founded arguments from cognitive theory (the SL analog would be a binary classifier distinguishing two class labels). It assumes that *cognitive subtraction* is possible, that is, the regional brain responses of interest can be isolated by constrasting two sets of brain scans that precisely differ in the cognitive facet of interest (Friston et al., 1996; Stark & Squire, 2001). For one voxel location at a time, an attempt is made to reject the null hypothesis of no difference between the averaged neural activity of a target brain state and the averaged neural activity of a control brain state. Note that establishing a brain-behavior association via GLM is therefore a judgment about neural activity compressed in the time dimension. The univariate GLM analysis can be easily extended to more than one output (dependent) variable within the CS regime by performing a multivariate analysis of covariance (MANCOVA). This allows for tests of more complex hypotheses but incurs multivariate normality assumptions (Nikolaus Kriegeskorte, 2011). Conceptually, one can switch the direction of inference by reframing the estimated beta coefficients as independent variables and the stimulus/task/behavior indicator as the response variable to perform multiple linear regression (ANCOVA) (Brodersen, Haiss, et al., 2011; T. Naselaris et al., 2011). The beta coefficients would then play an analogous role to the model weights of a trained classification algorithm. Finally, the often-performed *small volume correction* (an analog in the SL world would be classification after *feature selection*) corresponds to simultaneously fitting a GLM to each voxel of a restricted region of interest.

Because hypothesis testing for significant differences between beta coefficients of fitted GLMs relies on averaged neural activity, the test results are not corrupted by the conventionally applied spatial smoothing with a Gaussian filter. On the contrary, this even helps the correction for multiple comparisons based on random fields theory, alleviates inter-individual neuroanatomical variability, and thus increases sensitivity. Spatial smoothing however discards fine-grained spatial activity patterns that carry potential information about mental operations (J.-D. Haynes, 2015). Indeed, some authors believe that sensory, cognitive, and motor processes manifest themselves as neuronal population codes (Averbeck, Latham, & Pouget, 2006). Relevance of such population codes in human neuroimaging was for instance suggested by revealing subject-specific fusiform-gyrus responses to facial stimuli (Saygin et al., 2012). In applications of SL models, the spatial smoothing step is therefore often skipped because the "decoding" algorithms precisely exploit this multivariate structure of the salt-and-pepper patterns.

In contrast, decoding models use learning algorithms for an *informational agenda* by showing *generalization of robust patterns* to new brain activity acquisitions (de-Wit, Alexander, Ekroll, & Wagemans, 2016; N. Kriegeskorte et al., 2006; Mur et al., 2009). Information that is locally weak but spatially distributed can be effectively harvested in a structure-preserving fashion (J. D. Haynes & Rees, 2006). Some brain-behavior associations might only emerge when simultaneously capturing neural activity in a group of voxels but disappear in single-voxel approaches, such as GLMs. However, analogous to multivariate variants of the GLM, "decoding" can also be done by means of classical statistical approaches. Essentially, inference on information patterns in the brain reduces to model comparison (Friston, 2009). During training of a classifier to predict indicators correctly, an optimization algorithm (e.g., gradient descent or its variants) searches iteratively through the *hypothesis space* (= *function space*) of the chosen learning model. Each such hypothesis corresponds to one specific combination of model weights that equates with one candidate mapping function from the neural activity features to the indicators. In this way, four types of neuroscientific questions have been proposed to become quantifiable (Brodersen, 2009; Pereira et al., 2009): i) *Where* is an information category neurally processed? This extends the interpretational spectrum from increase and decrease of neural activity to the existence of complex combinations of activity variations distributed across voxels. For instance, linear classifiers decoded object categories from the ventral temporal cortex, even after excluding the fusiform gyrus, which is known to be responsive to object stimuli (Haxby et al., 2001). ii) *Whether* a given information category is reflected by neural activity? This extends the interpretational spectrum to topographically similar but neurally distinct processes that potentially underlie different cognitive facets. For instance, linear classifiers successfully whether a subject is attending to the first or second of two simultaneously presented gratings (Kamitani & Tong, 2005). iii) *When* is an information category generated (i.e., onset), processed (i.e., duration), and bound (i.e., alteration)? When applying classifiers to neural time series, the interpretational spectrum can be extended to the beginning, evolution, and end of distinct cognitive facets. For instance, different classifiers have been demonstrated to map the decodability time structure of mental operation sequences (King & Dehaene, 2014). iv) More controversially, *how* is an information category neurally processed? The interpretational spectrum is extended to computational properties of the neural processes, including processing in brain regions versus networks or isolated versus partially shared processing facets. For instance, a classifier trained for evolutionarily conserved eye gaze processes was able to decode evolutionarily more recent mathematical calculation processes as a possible case of neural recycling in the human brain (Knops, Thirion, Hubbard, Michel, & Dehaene, 2009). As an important caveat, the particular properties of a chosen learning algorithm (e.g., linear versus non-linear support vector machines) can probably not serve as a convincing argument for reverse-engineering neural processing mechanisms (Misaki, Kim, Bandettini, & Kriegeskorte, 2010). However, many prediction problems in neuroscience can probably be solved without exhaustive neurobiological micro-, meso-, or macro-level knowledge (H. Markram, 2006; Sandberg & Bostrom, 2008).

**4. Different theories**

**4.1 Diverging ways to formalize statistical inference**

Besides diverging historical origins and modelling goals, CS and SL rely on largely distinct theoretical frameworks that revolve around *null-hypothesis testing* and *statistical learning theory*. CS probably laid down its most important theoretical framework in the Popperian spirit of critical empiricism (Popper, 1935/2005): scientific progress is to be made by continuous replacement of current hypotheses by ever more pertinent hypotheses using *verification* and *falsification*. The rationale behind hypothesis falsification is that one counterexample can reject a theory by *deductive reasoning*, while any quantity of evidence cannot confirm a given theory by inductive reasoning (Goodman, 1999). The investigator verbalizes two mutually exclusive hypotheses by domain-informed judgment. The *alternative hypothesis* should be conceived so as to contradict the state of the art of the research topic. The *null hypothesis* should automatically deduce from the newly articulated alternative hypothesis. The investigator has the agenda to disprove the null hypothesis because it only leaves the preferred alternative hypothesis as the new standard belief. A conventional 5%-threshold (i.e., equating with roughly two standard deviations) guards against rejection due to the idiosyncrasies of the sample that are not representative of the general population. If the data have a probability of <=5% to be true given the null hypothesis ($P(result|H0)$), it is evaluated to be significant. Such a *test for statistical significance* indicates a difference between two means with a 5% chance after sampling twice from the same population. There are common misconceptions that it denotes the probability of the null hypothesis ($P(H0)$), the alternative hypothesis ($P(H1)$), the result of the test statistic ($P(result)$), or the null hypothesis given the result ($P(H0|result)$) (Markus, 2001; Pollard & Richardson, 1987). If the null hypothesis is not rejected, then nothing can be concluded about the whole significance test according to most statisticians. That is, the test yields no conclusive result, rather than a null result (Schmidt, 1996). In this way, classical hypothesis testing continuously replaces currently embraced hypotheses explaining a phenomenon in nature by better hypotheses with more empirical support in a Darwinian selection process. Note however that corroborating *substantive hypotheses* (e.g., a specific linguistic theory like Chomsky's Universal Grammar) requires more than statistical hypothesis testing (Chow, 1998; Friedman, 1998; Meehl, 1978). A statistical hypothesis can be properly tested in the absence of a substantive hypothesis of the phenomenon under study (Oakes, 1986). Finally, Fisher, Neyman, and Pearson intended hypothesis testing as a marker for further investigation, rather than an off-the-shelf decision-making instrument (J. Cohen, 1994; Nuzzo, 2014).

A theoretical framework relevant to both CS and SL, is the *bias-variance tradeoff*. It is a stand-alone concept that is helpful to consider in theory and practice by providing a different angle on deriving conclusions from data (e.g., Geman, Bienenstock, & Doursat, 1992). As famously noted[7], any statistical model is an over-simplification of the studied phenomenon, a distortion of the truth. Choosing the "right" statistical model hence pertains to the right balance between *bias* and *variance.* Bias denotes the difference between the *target function* (i.e., the "true" relationship between the data and a response variable that the investigator is trying to uncover from the data) and the average of the *function space* (or *hypothesis space*) instantiated by a model. Intuitively, the bias describes the rigidity of the model-derived functions. *High-bias models* yield virtually the same best approximating function, even for larger perturbations of the data. Variance denotes the difference between the best approximating function among members of the function space and the average of the function space. Intuitively, variance describes the differences between the functions derived from the same model family. *High-variance models* yield very different best approximating function even for small perturbations of the data. In sum, bias tends to decrease and variance tends to increase as *model complexity* increases. Simple models exhibit high bias and low variance. They lead to a bad *approximation* as they do not well simulate the target function, a behavior of the studied phenomenon. Yet, they exhibit a good *generalization* in successfully extrapolating from seen data to unseen data. Contrarily, complex models exhibit low bias and high variance. They have a better chance of well approximating the target function at the price of generalizing less well to new data.

---

[7] "All models are wrong; some models are useful." (George Box)

That is because complex models tend to fit too well the data to the point of fitting noise. This is called *overfitting* and occurs when model fitting integrates information independent of the target function and idiosyncratic to the data sample. The *bias-variance decomposition* captures this fundamental tradeoff in statistical modelling between approximating the behavior of the studied phenomenon and generalizing to new data describing that behavior. If the target function was known, the bias-variance tradeoff could be computed explicitly, in contrast to the inability of computing the Vapnik-Chervonenkis dimensions (VC dimensions) of non-trivial models (cf. next passage). The bias-variance tradeoff can also practically explain why successful applications of statistical models largely rely on i) the amount of available data, ii) the typically not known amount of noise in the data, and iii) the unknown complexity of the target function (Abu-Mostafa et al., 2012).

While the concept of *bias-variance tradeoff* is highly relevant in both abstract theory and everyday application of CS and SL models (Bishop, 2006; T. Hastie et al., 2001), the concept of *Vapnik-Chervonenkis dimensions* plays a crucial role in *statistical learning theory*. The VC dimensions mathematically formalize the circumstances under which a pattern-learning algorithm can successfully distinguish between points and extrapolate to new examples (Vapnik, 1989, 1996). This comprises any instance of learning from a number of observations to derive general rules that capture properties of phenomena in nature, including human and computer learning (cf. Bengio, 2014; Lake, Salakhutdinov, & Tenenbaum, 2015; Tenenbaum, Kemp, Griffiths, & Goodman, 2011). Note that this *inductive logic* to learn a general principle from examples contrasts the deductive logic of hypothesis falsification (cf. above). VC dimensions provide a probabilistic measure of whether a certain model is able to learn a distinction with respect to a given dataset. Formally, the VC dimensions measure the complexity capacity of a function space by counting the number of data points that can be cleanly divided (i.e., "shattered") into distinct groups as a result of the flexibility of the function set. Intuitively, the VC dimensions provide a guideline for the largest set of examples fed into a learning function such that it is possible to guarantee zero classification errors. They can be viewed as the effective number of parameters or the degrees of freedom, a concept shared between CS and SL. In this way, the VC dimensions formalize the circumstances under which a class of functions is able to learn from a finite amount of data to successfully predict a given phenomenon in previously unseen data. As one of the most important results from statistical learning theory, the number of configurations one can obtain from a classification algorithm grows polynomially, while the error is decreasing exponentially (Wasserman, 2013). Practically, good models have *finite VC dimensions* - fitting with a sufficiently large amount of data yieldes performances that approximate the theoretically expected performance in unseen data. Bad models have *infinite VC dimensions* - regardless of the available amount of data, it is impossible to make generalization conclusions on unseen data. In contrast to the bias-variance tradeoff, the VC dimensions (like null-hypothesis testing) are unrelated to the *target function,* as the "true" mechanisms underlying the studied phenomenon in nature. Yet, VC dimensions relate to the models used to approximate that target function. As an interesting consideration, it is possible that generalization from a concrete dataset fails even if the VC dimensions predict learning as very likely. However, such a dataset is theoretically unlikely to occur. Although the VC dimensions are the best formal concept to derive errors bounds in statistical learning theory (Abu-Mostafa et al., 2012), they can only be explicitly computed for simple models. Hence, investigators are often restricted to an approximate bound for VC dimensions, which limits their usefulness for theoretical considerations.

**4.2 The impact of diverging inference categories on neuroimaging analyses**

When looking at neuroimaging research through the CS lens, statistical estimation revolves around solving the *multiple comparisons problem* (Thomas E. Nichols, 2012; T.E. Nichols & Hayasaka, 2003). From the SL stance, however, it is the *curse of dimensionality* and *overfitting* that statistical analyses need to tackle (Domingos, 2012; Friston et al., 2008). In typical neuroimaging studies, CS methods typically test one hypothesis many times (i.e., the null hypothesis), whereas SL methods typically search through thousands of different hypotheses in a single process (i.e., walking through the function space by numerical optimization). The high voxel resolution of common brain scans offers parallel measurements of >100,000 brain locations. In a mass-univariate regime, such as after fitting voxel-wise GLMs, the same statistical test is applied >100,000 times. The more often the investigator tests a hypothesis of relevance for a brain location, the more locations will be falsely detected as relevant (false positive, Type I error), especially in the noisy neuroimaging data. The issue consists in too many simultaneous statistical inferences. From a general perspective, all dimensions in the data (i.e., voxel variables) are implicitly treated as equally important and no neighborhoods of most expected variation are statistically exploited (T. Hastie et al., 2001). Hence, the absence of complexity restrictions during the statistical modelling of neuroimaging data takes a heavy toll at the final inference step.

This is contrasted by the high-dimensional SL regime, where the initial model choice by the investigator determines the complexity restrictions to all data dimensions (i.e., not single voxels) that are imposed explicitly or implicitly by the model structure. Model choice predisposes existing but unknown low-dimensional neighborhoods in the full voxel space to achieve the prediction task. Here, the toll is taken at the beginning because there are so many different alternative model choices that would impose a different set of complexity constraints. For instance, signals from "brain regions" are likely to be well approximated by models that impose discrete, locally constant compartments on the data (e.g., k-means or spatially constrained Ward clustering). Tuning model choice to signals from macroscopical "brain networks" should impose overlapping, locally continuous data compartments (e.g., independent component analysis or sparse principal component analysis). Knowledge of such *effective dimensions* in the neuroimaging data is a rare opportunity to simultaneously reduce the model bias *and* model variance, despite their typically inverse relationship. Statistical models that overcome the curse of dimensionality typically incorporate an explicit or implicit metric for such anisotropic neighborhoods in the data (Bach, 2014; Bzdok, Eickenberg, Grisel, Thirion, & Varoquaux, 2015; T. Hastie et al., 2001). Viewed from the bias-variance tradeoff, this successfully calibrates the sweet spot between underfitting and overfitting. Viewed from statistical learning theory, the VC dimensions can be reduced and thus the generalization performance increased. Applying a model without such complexity restrictions to high-dimensional brain data, generalization becomes difficult to impossible because all directions in the data are treated equally in with isotropic structure. At the root of the problem, all data samples look virtually identical in high-dimensional data scenarios ("curse of dimensionality", Bellman, 1961). The learning algorithm will not be able to see through the noise and will thus overfit. In fact, these considerations explain why the multiple comparisons problem is closely linked to encoding studies and overfitting is more closely related to decoding studies (Friston et al., 2008). Moreover, it offers explanations as to why analyzing neural activity in a region of interest, rather than the whole brain, simultaneously alleviates both the multiple comparisons problem (called "small volume correction" in CS studies) and the overfitting problem (called "feature selection" in SL studies).

Further, some common invalidations of the CS and SL statistical frameworks are conceptually related (cf. case study two). An often-raised concern in neuroimaging studies performing classical inference is *double dipping* or *circular analysis* (N. Kriegeskorte et al., 2009). This occurs when, for instance, first correlating a behavioral measure with brain activity and then using the identified subset of brain voxels for a second correlation analysis with that same behavioral measurement (Lieberman, Berkman, & Wager, 2009; Vul et al., 2008). In this scenario, voxels are submitted to two statistical tests with the same goal in a nested, non-independent fashion[8]. This corrupts the *validity of the null hypothesis* on which the reported test results conditionally depend. Importantly, this case of repeating a same statistical estimation with iteratively pruned data selections (on the training data split) is a valid routine in the SL framework, such as in recursive feature extraction (Isabelle Guyon, Weston, Barnhill, & Vapnik, 2002; Hanson & Halchenko, 2008). However, there is an analog to double-dipping or circular analysis in CS methods applied to neuroimaging data: *data-snooping* or *peeking* in SL analyses based on out-of-sample generalization (Abu-Mostafa et al., 2012; Fithian, Sun, & Taylor, 2014; Pereira et al., 2009). This occurs, for instance, when performing simple (e.g., mean-centering) or more involved (e.g., k-means clustering) target-variable-dependent or -independent preprocessing on the entire dataset, where it should be applied separately to the training sets and test sets. Data-snooping can lead to optimistic cross-validation estimates and a trained learning algorithm that fails on fresh data drawn from the same distribution. Rather than a corrupted null hypothesis, it is the *error bounds of the VC dimensions that are loosened* and, ultimately, invalidated because information from the concealed test set influences model selection on the training set. Conceptually, these two distinct classes of statistical errors arising within the CS and SL frameworks have several common points: i) Both involve a form of information compression by the necessity to draw conclusions on small data parts drawn from originally high-dimensional neuroimaging data. ii) Unauthorized prior knowledge happens to be introduced into the statistical estimation process. iii) Both *double dipping* (CS) and *data snooping* (SL) involve a form of *bias* - biasing the estimates to systematically deviate from the population parameter or increasing the bias error term in the bias-variance decomposition (Tal Yarkoni & Westfall, 2016). iv) Both types of faux pas in statistical conduct can occur in very subtle and unexpected ways, but can be avoided with often small effort (see case study two). v) Invalidation of the null hypothesis or the VC bounds can yield overly optimistic results that may encourage unjustified confidence in neuroscientific findings and conclusions.

---

[8] "If you torture the data enough, nature will always confess." (Ronald Coase)

Moreover, it is probably optimal to perform cross-validation with k = 5 or 10 data splits in neuroimaging, despite diverging choices in previous studies. This is because empirical simulation studies (Leo Breiman & Spector, 1992; Kohavi, 1995) have indicated that more data splits k (i.e., bigger training set, smaller test set) can increase the estimate's variance. Fewer data splits k (i.e., smaller training set, bigger test set) can increase the estimate's bias. Concretely, for a particular estimate issued by leave-one-out cross-validation (k data splits = n samples) the neuroimaging practioner might unluckily obtain a "fluctuation" far from the ground-truth value and there is often no second dataset to double-check. As a drastic sidenote, neuroscientists do not even know whether a target function exists in nature when approximating functions when, for instance, automatically classifying mental operations or healthy from diseased individual based on brain scans (Nesterov, 2004; Wolpert, 1996). Additionally, an analytical proof (Shao, 1993) showed that leave-one-out cross-validation does not provide a theoretical guarantee for consistent model estimation. That is, this cross-validation scheme does not always select the best model assuming that it is known. However, the neuroimaging practitioner is typically satisfied with model weights that are reasonably near to an optimal local point and does not focus on reaching the single global optimal point as typical in convex optimization (Boyd & Vandenberghe, 2004). As an exception, leave-one out cross-validation may therefore be more justified in neuroimaging data scenarios with particularly few samples, where a decent model fit is the primary concern. On a more practical note, the computational load for leave-one-out cross-validation with k=n is often much higher than for k-fold cross-validation with k=5 or 10.

**5. Different currencies**

**5.1 Diverging performance metrics can quantify the behavior of statistical models**

The neuroscientific investigator who adopts a CS culture, typically somebody with a background in psychology, biology, or medicine, is in the habit of diagnosing statistical investigations by means of p values, effect sizes, confidence intervals, and statistical power. The *p value* denotes the conditional probability of obtaining an equal or more extreme test statistic provided that the null hypothesis $H_0$ is true at the prespecified significance threshold alpha (Anderson, Burnham, & Thompson, 2000). Under the condition of sufficiently high power (cf. below), it quantifies the strength of evidence against the null hypothesis as a continuous function (Rosnow & Rosenthal, 1989). Counterintuitively, it is not an immediate judgment on the alternative hypothesis $H_1$ preferred by the investigator (Anderson et al., 2000; J. Cohen, 1994). Three main interpretations for the significance level exist (Gerd Gigerenzer, 1993): i) the *conventional* level of significance specified before the investigation yielding yes-no information (early Fisher), ii) the *exact* level of significance obtained after the investigation yielding continuous information (late Fisher), and iii) the *alpha level* indicating a Type I error frequency in tests after repeated sampling (Neyman/Pearson). The former views the obtained p value as a property of the data, whereas the latter views alpha as a property of the statistical test (Gerd Gigerenzer, 1993). P values do not qualify the possibility of replication. It is another important caveat that p values become better (i.e., lower) with increasing sample sizes (Berkson, 1938).

The essentially binary p value is therefore often complemented by the continuous *effect size*. The p value is a deductive *inferential* measure, whereas the effect size is a *descriptive* measure that follows neither inductive nor deductive reasoning. The effect size can be viewed as the strength of a statistical relationship, how much $H_0$ deviates from $H_1$, or the likely presence of an effect in the general population (Chow, 1998; Ferguson, 2009; Kelley & Preacher, 2012). It is a unit-free, sample-size independent, often standardized statistical measure for the importance of rejecting $H_0$. It equates to zero if H₀ could not be rejected. As a tendency, the lower the p value, the higher the effect size (Nickerson, 2000). Importantly, the effect size allows the identification of marginal effects that pass the statistical significance threshold but are not practically relevant in the real world. As a property of the actual statistical test, the effect size has different names and takes various forms, such as *rho* in Pearson correlation, *eta²* in explained variances, and *Cohen's d* in differences between group averages.

Additionally, the certainty of a *point estimate* (i.e., the outcome is a value) can always be expressed by an *interval estimate* (i.e., the outcome is a value range) using *confidence intervals.* They indicate with a chosen probability how often the "true" population effect would be within the investigator-specified interval after many repetitions of the study (Cumming, 2009; Estes, 1997; Nickerson, 2000). Typically, a 95% confidence interval is spanned around the range of values of a sample mean statistic that includes the population mean in 19 out of 20 cases across all samples. The tighter the confidence interval, the smaller the variance of the point estimate of the population parameter in each draw sample (keeping the sample size constant). The sometimes normalized confidence intervals can be computed in a variety of ways. Their estimation is influenced by sample size and population variability. They can be reported for different statistics, with different percentage borders, and may be asymmetrical. Note that confidence intervals can be used as a viable surrogate for formal tests of statistical significance in many scenarios (Cumming, 2009).

Confidence intervals can be computed in various data scenarios and statistical regimes, whereas the *power* is only meaningful within the culture of formal hypothesis falsification (Jacob Cohen, 1977, 1992; Oakes, 1986). The quality of interpretation for a statistically significant result is strongly affected by whether the original hypothesis was tenable. The power measures the probability of a statistical test to find a "true" effect, of rejecting H₀ in the long term, or how well a "true" alternative hypothesis is correctly accepted assuming the effect exists in the population, that is, P(H₀ detected to be false|H₁ true). A high power thus ensures that statistically significant and non-significant tests indeed reflect a property of the population (Chow, 1998). Intuitively, small confidence intervals are an indicator of high statistical power. Type II errors (i.e., false negatives, beta error) become less likely with higher power (= 1 - beta error). Concretely, an underpowered investigation does not allow choosing between $H_0$ and $H_1$ at the specified significance threshold alpha. Power calculations depend on several factors, including significance threshold alpha, the effect size in the population, variation in the population, sample size n, and experimental design (Jacob Cohen, 1992). Rather than retrospective, the necessary sample size n for a desired power can be computed in a prospective fashion after specifying an alpha and a hypothesized effect size.

In contrast, diagnosis of the obtained research findings takes a different shape for the SL-indoctrinated neuroscientist[9], typically somebody with a background in computer science, physics, or engineering. *Cross-validation* is the de facto standard to obtain an unbiased estimate of a model's capacity to generalize beyond the sample at hand (Bishop, 2006; T. Hastie et al., 2001). *Model assessment* is done by training on a bigger subset of the available data (i.e., *training set* for *in-sample performance*) and subsequent application of the trained model to the smaller remaining part of data (i.e., *test set* for *out-of-sample performance*), which is assumed to share the same distribution. Cross-validation thus permutes over the sample in data splits until the class label (i.e., categorical target variable) of each data point has been predicted once. This set of model-predicted labels and the corresponding true data point labels can then be submitted to the quality measures *accuracy*, *precision*, *recall*, and *F1 score* (Powers, 2011). As the simplest among them, *accuracy* is a summary statistic that captures the fraction of correct prediction instances among all performed model applications. This and the following measures are often computed separately on the training set and the test set. Additionally, the measures from training and testing can be expressed by their inverse (e.g., *training error* as *in-sample error* and *test error* as *out-of-sample error*) because the positive and negative cases are interchangeable.

---

[9] "It is also important to be particularly careful in reporting errors and measures of model fit in the high-dimensional setting. We have seen that when p > n, it is easy to obtain a useless model that has zero residuals. Therefore, one should *never* use sum of squared errors, p-values, $R^2$ statistics, or other traditional measures of model fit on the training data as evidence of a good model fit in the high-dimensional setting." (James, Witten, Hastie, & Tibshirani, 2013, p. 247, authors' emphasis)

The *classification accuracy* (= 1 - classification error) can be further decomposed into class-wise metrics based on the so-called *confusion matrix*, the juxtaposition of the true and predicted class memberships. The *precision* (= true positive / (true positive + false positive)) measures how many of the predicted labels are correct, that is, how many members predicted to belong to a class really belong to that class. For instance, among the participants predicted to have depression, how many are really affected by that disease? On the other hand, the *recall* (= true positive / (true positive + false negative)) measures how many labels are correctly predicted, that is, how many members of a class were predicted to really belong to that class. Hence, among the participants affected by depression, how many were actually detected as such? Put differently, precision can be viewed as a measure of "exactness" or "quality" and recall as a measure of "completeness" or "quantity" (Powers, 2011). Neither accuracy, precision, or recall allow injecting subjective importance into the evaluation process of the prediction model. This disadvantage is alleviated by the *$F_{beta}$ score*, which is a weighted average of the precision and recall prediction scores. Concretely, the $F_1$ score would equally weigh precision and recall of class predictions, while the $F_{0.5}$ score puts more emphasis on precision and the $F_2$ score more on recall. Moreover, applications of recall, precision, and $F_{beta}$ scores have been noted to ignore the true negative cases as well as to be highly susceptible to estimator bias (Powers, 2011). Needless to say, no single measure can be equally optimal in all contexts.

Finally, *learning curves* (Abu-Mostafa et al., 2012; Murphy, 2012) are an important diagnostic tool to evaluate *sample complexity*, that is, the achieved model fit and prediction as a function of the available sample size n. For increasingly bigger subsets of the training set, a classification algorithm is trained on that current share of the training set and then evaluated for accuracy on the always-same test set. Across subset instances, simple models display relatively high in-sample error because they cannot approximate the target function very well (underfitting) but exhibit good generalization to unseen data with relatively low out-of-sample error. Conversely, complex models display relatively low in-sample error because they adapt too well to the data (overfitting) with difficulty to extrapolate to newly sampled data with high out-of-sample error. In both scenarios, the model effectiveness is more "challenged", the scarcer the data points available for model training.

**5.2 Outcome metrics of statistical models used in neuroimaging**

Reports of statistical outcomes in the neuroimaging literature have previously been recognized to confuse notions from classical statistics and statistical learning (Friston, 2012). On a general basis, *CS and SL do not judge findings by the same aspects of evidence* (Lo et al., 2015; Shmueli, 2010). In neuroimaging papers based on classical hypothesis-driven inference, p values, and confidence intervals are ubiquitously reported. There have however been very few reports of effect size in the neuroimaging literature (N. Kriegeskorte, Lindquist, Nichols, Poldrack, & Vul, 2010). Effect sizes, in turn, are necessary to compute power estimates. This explains the even rarer occurrence of power calculations in neuroimaging research (but see Tal Yarkoni & Braver, 2010). Effect sizes have been argued to allow verification that in an optimal experimental design inference is tuned to big effect sizes (Friston, 2012). To, for instance, estimate the p value and the effect size for local changes in neural activity during a psychological task, one would actually need two independent samples of these experimental data. One sample would be used to perform statistical inference on the neural activity change and one sample to obtain unbiased effect sizes. It has been previously emphasized (Friston, 2012) that p values and effect sizes reflect in-sample estimates in a retrospective inference regime (CS). These metrics find an analogue in out-of-sample estimates issued from cross-validation in a prospective prediction regime (SL). In-sample effect sizes are typically an *optimistic* estimate of the "true" effect size (inflated by high significance thresholds), whereas out-of-sample effect sizes are *unbiased* estimates of the "true" effect size. As an important consequence, neuroimaging investigators should refrain from simultaneously computing and reporting both types of estimates on an identical data sample. This can lead to double dipping (cf. case study two).

In the high-dimensional scenario, analyzing "wide" neuroimaging data in our case, judging statistical significance by p values is generally considered to be challenging (Bühlmann & Van De Geer, 2011;

see case studies two and three)**.** Instead, classification accuracy on fresh data is probably the most often-reported performance metric in neuroimaging studies using learning algorithms. Basing interpretation on accuracy alone is however influenced by the local characteristics of hemodynamic responses, efficiency of experimental design, data folding into train and test sets, and differences in the feature number p (J.-D. Haynes, 2015). A potentially under-exploited SL tool in this context is *bootstrapping.* It enables population-level inference of unknown distributions independent of model complexity by repeated random draws from the neuroimaging data sample at hand (Efron, 1979; Efron & Tibshirani, 1994). This opportunity to equip various point estimates by an interval estimate of certainty (e.g., the interval for the "true" accuracy of a classifier) is unfortunately seldom embraced in the contemporary neuroimaging domain (but see Bellec, Rosa-Neto, Lyttelton, Benali, & Evans, 2010; Vogelstein et al., 2014). Besides providing confidence intervals, bootstrapping can also perform non-parametric null hypothesis testing. This may be a rare example of a direct connection between CS and SL methodology. Alternatively, *binomial tests* can be used to obtain a p-value estimate of statistical significance from accuracies and other performance scores (Brodersen et al., 2013; Hanke, Halchenko, & Oosterhof, 2015; Pereira et al., 2009) in the binary classification setting (for an example see Bludau et al., 2015). It can reject the null hypothesis that two categories occur equally often. A last option that is also applicable to the multi-class setting is *label permutation*, another non-parametric resampling procedure (Golland & Fischl, 2003; T. E. Nichols & Holmes, 2002). It can serve to reject the null hypothesis that the neuroimaging data do not contain any information about the class labels.

Extending from the setting of two hypotheses or yes-no classification to multiple classes injects ambiguity into the interpretation of accuracy scores. Rather than mere better-than-chance findings, it becomes more important to evalute the $F_1$, precision and recall scores for each class to be predicted in the brain scans (e.g., Brodersen, Schofield, et al., 2011; Schwartz, Thirion, & Varoquaux, 2013). It is important to appreciate that the sensitivity/specificity metrics, more frequently reported in CS communities, and the precision/recall metrics, more frequently reported in SL communities, tell slightly different stories about identical neuroscientific findings. In fact, sensitivity equates with recall (= true positive / (true positive + false negative)). Specificity (= true negative / (true negative + false positive)) does however not equate with precision (=true positive / (true positive + false positive)). Further, a CS view on the SL metrics would be that maximum precision corresponds to absent Type I errors (i.e., no false positives), whereas maximum recall corresponds to absent Type II errors (i.e., no false negatives). Again, Type I and II errors are related to the entirety of datapoints in a CS regime and prediction is only evaluated on a test data split of the sample in a SL regime. In learning curves, a big gap between high in-sample and low out-of-sample performance is typically observed for high-variance models, such as neural network algorithms or random forests. These performance metrics from different data splits often converge for high-bias models, such as linear support vector machines and logistic regression. Moreover, the medical domain and social sciences usually aggregate results in *ROC* (receiver operating characteristic) curves plotting sensitivity against specificity scores, whereas engineering and computer science domains tend to report *recall-precision curves* instead (Davis & Goadrich, 2006; Demšar, 2006).

Finally, it is often possible to inspect the *fit for purpose* of trained black-box models (Brodersen, Haiss, et al., 2011; Kuhn & Johnson, 2013). In SL this can take the form of evaluating some notion of *support recovery*, that is, the question to what extent the learning algorithms put probability mass on the parts of the feature space that are "truely" underlying the given classes. In neuroimaging, this pertains to the difference between models that capture task-specific aspects of neural activity or arbitrary discriminative aspects, such as structured noise in participant- or scanner-related idiosyncracies (Brodersen, Haiss, et al., 2011; Varoquaux & Thirion, 2014). Such a face-validity criterion of meaningful model fit is all the more important because of the general trade-off in statistics between choosing models with best possible performance and those with model parameters that are most interpretable (T. Hastie et al., 2001). Instead of maximizing prediction scores, neuroimaging investigators might want to focus on *neurobiologically informed feature spaces* and *mechanistically interpretable model weights* (Brodersen, Schofield, et al., 2011; Bzdok et al., 2015; K. E. Stephan, 2004). In fact, SL neuroimaging studies would perhaps benefit from a metric for neurobiological plausibilty as an acid test during model selection. Reverse-engineering fitted models by *reconstruction* of stimulus or task aspects (Miyawaki et al., 2008; Thomas Naselaris, Prenger, Kay, Oliver, & Gallant, 2009; Thirion et al., 2006) could become an important evaluation metric for *why* a learned feature-label mapping exihibits a certain model performance (Gabrieli et al., 2015). This may further disambiguate explanations for (statistically significant) better-than-chance accuracies because i) numerous, largely different members of a given function space may yield essentially similar model performance and ii) each fitted model may only capture a part of the class-relevant structure in the neuroimaging data (cf. Nikolaus Kriegeskorte, 2011). For instance, fit-for-purpose metrics used in neuroimaging could uncover what neurobiological aspects allow classifying an individual as schizophrenic versus normal and which neurobiological endophenotypes underlie the schizophrenia "spectrum" (Bzdok & Eickhoff, 2015; Hyman, 2007; Klaas E. Stephan et al., 2015; K. E. Stephan, Friston, & Frith, 2009).

**6. Case study one: Generalization and subsequent classical inference**

Vignette: We are interested in potential differences in brain structure that are associated with an individual's age (*continuous target variable*). A Lasso (*belongs to SL arsenal*) is computed on the voxel-based morphometry data (Ashburner & Friston, 2000) from the brain's grey matter of the 500-subject HCP release (Human Connectome Project; Van Essen et al., 2012). This L1-penalized residual-sum-of-squares regression performs variable selection (i.e., *effectively eliminates coefficients by setting them to zero*) on all grey-matter voxels' volume information in a *high-dimensional* (not mass-univariate) regime. Assessing *generalization* performance of different sparse models using five-fold cross validation yields the non-zero coefficients for few brain voxels whose volumetric information is most *predictive* of an individual's age.

Question: How can we perform *classical inference* to known which of the grey-matter voxels selected to be predictive for biological age are *statistically significant*?

This is an important concern because most statistical methods currently applied to large datasets perform some explicit or implicit form of variable selection (Trevor Hastie et al., 2015; Jenatton et al., 2011; Jordan et al., 2013). There are even many different forms of preliminary selection of variables before performing significance tests on them. First, Lasso is a widely used estimator in engineering, compressive sensing, various "omics" branches, and other sciences, mostly without a significance test. Beyond neuroscience, generalization-approved statistical learning models are routinely solving a diverse set of real-world challenges. This includes but is not limited to algorithmic trading in financial markets, real-time speech translation, SPAM filtering for e-mails, face recognition in digital cameras, and piloting self-driving cars (Jordan & Mitchell, 2015; LeCun et al., 2015). In all these examples statistical learning algorithms successfully generalize to unseen data and thus tackle the problem heuristically without classical significance test for variables or model performance.

Second, the Lasso has solved the combinatorial problem of what subset of grey-matter voxels best predicts an individual's age by *automatic variable selection*. Computing voxel-wise p values would recast this high-dimensional pattern-learning setting into a mass-univariate hypothesis-testing problem where relevance would be computed independently for each voxel and correction for multiple comparisons would become necessary. Yet, recasting into the mass-univariate setting would ignore the sophisticated selection process that led to the predictive model with a reduced number of variables (Wu et al., 2009). Put differently, the variable selection procedure is itself a stochastic process that is however not accounted for by the theoretical guarantees of classical inference for statistical significance (Berk, Brown, Buja, Zhang, & Zhao, 2013). Put in yet another way, data-driven model selection is corrupting hypothesis-driven statistical inference because the sampling distribution of the parameter estimates is altered. The important consequence is that naive classical inference expects a non-adaptive model chosen before data acquisition and can therefore not be used along Lasso in particular or arbitrary selection procedures in general[10].

Third, this conflict between data-guided model selection by cross-validation (SL) and confirmatory classical inference (CS) is currently at the frontier of statistical development (Loftus, 2015; J. Taylor & Tibshirani, 2015). New methods for so-called *post-selection inference* (or *selective inference*) allow computing p values for a set of features that have previously been chosen to be meaningful predictors by some criterion. According to the theory of CS, the statistical model is to be chosen before visiting the data. Classical statistical tests and confidence intervals therefore become invalid and the p values become downward-biased (Berk et al., 2013). Consequently, the association between a predictor and the target variable must be even stronger to certify on the same level of significance. Selective inference for modern adaptive regression thus replaces loose *naive p values* by more rigorous *selection-adjusted p values*. As an ordinary null hypothesis can hardly be adopted in this adaptive testing setting, conceptual extension is also prompted on the level of CS theory itself (Trevor Hastie et al., 2015). Closed-form solutions to adjusted inference after variable selection already exist for principal component analysis (Choi, Taylor, & Tibshirani, 2014) and forward stepwise regression (Jonathan Taylor, Lockhart, Tibshirani, & Tibshirani, 2014). Last but not least, a simple

---

[10] "Once applied only to the selected few, the interpretation of the usual measures of uncertainty do not remain intact directly, unless properly adjusted." (Yoav Benjamini)

alternative to formally account for preceeding model selection is *data splitting* or *sample splitting* (D. R. Cox, 1975; Fithian et al., 2014; Wasserman & Roeder, 2009), which is frequent practice in genetics (e.g., Sladek et al., 2007). In this procedure, the selection procedure is computed on one data split and p values are computed on the remaining second data split. However, data splitting is not always possible and will incur power losses and interpretation problems.

**7. Case study two: Classical inference and subsequent generalization**

Vignette: We are interested in potential brain structure differences that are associated with an individual's gender (*categorical target variable*) in the voxel-based morphometry data (Ashburner & Friston, 2000) of the 500-subject HCP release (Human Connectome Project; Van Essen et al., 2012). Initially, the >100,000 voxels per brain scan are reduced to the most important 10,000 voxels to lower the computational cost and facilitate predictive model estimation. To this end, ANOVA (*univariate test for statistical significance belonging to CS*) is first used to obtain a ranking of the most relevant 10,000 features from the grey matter of each subject. This selects the 10,000 out of the original >100,000 voxel variables with highest variance explaining volume differences between males and females (i.e., *the male-female class labels are used in the univariate test*). Second, support vector machine classification ("*multivariate" pattern-learning algorithm belonging to SL*) is performed by training and testing on a feature space with the 10,000 preselected grey-matter measurements to predict the gender from each subject's brain scan.

Question: Is an analysis pipeline with *univariate classical inference* and subsequent *high-dimensional prediction* valid if both steps rely on the same target variables?

The implications of feature engineering procedures applied before training a learning algorithm is a frequent concern and can have very subtle answers (I. Guyon & Elisseeff, 2003; Hanke et al., 2015; N. Kriegeskorte et al., 2009; Lemm, Blankertz, Dickhaus, & Muller, 2011). In most applications of predictive models the large majority of brain voxels will be uninformative (Brodersen, Haiss, et al., 2011). The described scenario of *dimensionality reduction* by feature selection to focus prediction is clearly allowed under the condition that the ANOVA is not computed on the entire data sample. Rather, the voxels explaining most variance between the male and female individuals should be computed only on the training set in each cross-validation fold. In the training set and test set of each fold the same identified candidate voxels are then regrouped into a feature space that is fed into the support vector machine algorithm. This ensures an identical feature space for model training and model testing but its construction only depends on structural brain scans from the training set. Generally, any voxel preprocessing prior to model training is authorized if the feature space construction is not influenced by properties of the concealed test set. In the present scenario, the Vapnik-Chervonenkis bounds of the cross-validation estimator are therefore not loosened or invalidated if class labels have been exploited for feature selection or depending on whether the feature selection procedure is univariate or multivariate. Put differently, the cross-validation procedure simply evaluates the entire prediction process including the automatized and potentially nested dimensionality reduction procedure. In sum, in a SL regime, using class information during feature preprocessing for a cross-validated supervised estimator is not an instance of *data-snooping* (or *peeking*) if done exclusively on the training set (Abu-Mostafa et al., 2012).

This is an advantage of cross-validation yielding *out-of-sample estimates*. In stark contrast, remember that null-hypothesis testing yields *in-sample estimates*. Using the class labels for a variable selection step just before statistical hypothesis testing on a same data sample would invalidate the null hypothesis (N. Kriegeskorte et al., 2010; N. Kriegeskorte et al., 2009) (cf. case study one). Consequently, in a CS regime, using class information to select variables before null-hypothesis testing will incur an instance of *double-dipping* (or *circular analysis*).

Regarding interpretation of the results, the classifier will miss some brain voxels that only carry relevant information when considered in voxel ensembles. This is because the ANOVA filter kept voxels that are independently relevant (Brodersen, Haiss, et al., 2011). Univariate feature selection may systematically encourage model selection (i.e., each weight combination equates with a model hypothesis from the classifier's function space) that are not neurobiologically meaningful. Concretely, in the discussed scenario the classifier learns *complex patterns between voxels that were chosen to be individually important*. This may considerably weaken the interpretability and conclusions on "whole-brain multivariate patterns". Remember also that variables that have a *statistically significant association* with a target variable do not necessarily have good *generalization performance,* and vice versa (Lo et al., 2015; Shmueli, 2010). On the upside, it is widely believed that the combination of whole-brain univariate feature selection and linear classification is frequently among the best approaches if the primary goal is *optimized prediction performance* as opposed to *optimized interpretability*. Finally, it is interesting to consider that ANOVA-mediated feature selection of p < 500 voxel variables reduces the "wide" neuroimaging data ("n << p" setting) down to "long" neuroimaging data with fewer features than observations ("n > p" setting) given the n = 500 subjects (Wainwright, 2014). This allows recasting the SL regime into a CS regime in order to fit a standard general linear model and perform classical null-hypothesis testing instead of training a predictive classification algorithm (Brodersen, Haiss, et al., 2011).

## 8. Case study three: Structure discovery by clustering algorithms

Vignette: Each functionally specialized region in the human brain probably has a unique set of long-range connections (Passingham, Stephan, & Kotter, 2002). This notion has prompted connectivity-based parcellation methods in neuroimaging that segregate a region of interest (ROI, can be locally circumscribed or brain global; S. B. Eickhoff, Thirion, Varoquaux, & Bzdok, 2015) into distinct cortical modules (Behrens et al., 2003). The whole-brain connectivity for each ROI voxel is computed and the voxel-wise connectional fingerprints are submitted to a clustering algorithm (i.e., *individual brain voxels in the ROI are the elements to group; the connectivity strength values are the features of each element for similarity assessment*). In this way, connectivity-based parcellation issues cortical modules in the specified ROI that exhibit similar connectivity patterns and are, thus potentially, functionally distinct. That is, voxels within the same cluster in the ROI will have more similar connectivity properties than voxels from different ROI clusters.

Question: Is it possible to decide whether the obtained brain *clusters* are *statistically significant*?

Essentially, the aim of connectivity-guided brain parcellation is to find useful, simplified structure by imposing discrete compartments on brain topography (Frackowiak & Markram, 2015; S. M. Smith et al., 2013; Yeo et al., 2011). This is typically achieved by k-means, hierarchical, Ward, or spectral clustering algorithms (Thirion, Varoquaux, Dohmatob, & Poline, 2014). Putting on the CS hat, a ROI clustering result would be deemed statistically significant if it has a very low probability of being "true" under the null hypothesis that the investigator seeks to reject (Everitt, 1979; Halkidi, Batistakis, & Vazirgiannis, 2001). Choosing a test statistic for clustering solutions to obtain p values is difficult (Vogelstein et al., 2014) because of the need for a meaningful null hypothesis to test against (Jain, Murty, & Flynn, 1999). Put differently, for hypothesis-driven statistical inference one may need to pick an arbitrary hypothesis to falsify. It follows that the CS notions of effect size and power do not seem to apply in the case of brain parcellation. Instead of classical inference to formally *test* for a particular structure in the clustering results, we actually need to resort to exploratory approaches that *discover* and *assess* structure in the neuroimaging data (Efron & Tibshirani, 1991; T. Hastie et al., 2001; Tukey, 1962). Although statistical methods span a continuum between the two poles of CS and SL, finding a clustering model with the highest fit in the sense of explaining the regional connectivity differences at hand is more naturally situated in the SL community.

Putting on the SL hat, we realize that the problem of brain parcellation constitutes an unsupervised learning setting without any target variable y to predict (e.g., cognitive tasks, the age or gender of the participants). The learning task is therefore not to estimate a supervised predictive model y = f(X), but to estimate an unsupervised descriptive model for the connectivity data X themselves. Solving such unsupervised estimation problems is generally recognized to be very hard (Bishop, 2006; Zoubin Ghahramani, 2004; T. Hastie et al., 2001). In clustering problems, there are many possible transformations, projections, and compressions of X but there is no criterion of optimality that clearly suggests itself. On the one hand, the "true" *shape of clusters* is unknown for most real-world clustering problems, including brain parcellation studies. On the other hand, finding an "optimal" *number of clusters* represents an unresolved issue (*cluster validity problem*) in statistics in general and in brain neuroimaging in particular (Handl, Knowles, & Kell, 2005; Jain et al., 1999). Evaluating the model fit of clustering results is conventionally addressed by heuristic *cluster validity criteria* (S. B. Eickhoff et al., 2015; Thirion et al., 2014). These are necessary because clustering algorithms will always find subregions in the investigator's ROI, that is, relevant structure according to the clustering algorithm's optimization objective, whether these truly exist in nature or not. There is a variety of such criteria based on information theory, topology, and consistency. They commonly encourage cluster solutions with low within-cluster and high between-cluster differences, regardless of the applied clustering algorithm.

Evidently, the discovered connectivity clusters are mere hints to candidate brain modules. Their "existence" in neurobiology requires further scrutiny (S. B. Eickhoff et al., 2015; Thirion et al., 2014). Nevertheless, such clustering solutions are an important means to narrow down high-dimensional neuroimaging data. Preliminary clustering results broaden the space of research hypotheses that the investigator can articulate. For instance, unexpected discovery of a candidate brain region (cf. Mars et al., 2012; zu Eulenburg, Caspers, Roski, & Eickhoff, 2012) can provide an argument for future experimental investigations. Brain parcellation can thus be viewed as an exploratory unsupervised method outlining relevant structure in neuroimaging data that can subsequently be formally tested as a concrete hypothesis in neuroimaging studies whose interpretations are based on classical inference.

## 9. Conclusion

A novel scientific fact about the brain is only valid in the context of the complexity restrictions that have been imposed on the studied phenomenon during the investigation (Box, 1976). The statistical arsenal of the imaging neuroscientist can be divided into classical inference by hypothesis falsification and increasingly used generalization inference by extrapolating complex patterns. While null-hypothesis testing has been dominating the academic milieu for several decades, statistical learning methods are prevalent in many data-intense branches of industry (Vanderplas, 2013). This sociological segregation may partly explain existing confusion about their mutual relationship. Based on diverging historical trajectories and theoretical foundations, both statistical cultures aim at extracting new knowledge from data using mathematical models (Friston et al., 2008; Jordan et al., 2013). However, an observed effect with a statistically significant p value does not necessarily generalize to future data samples. Conversely, an effect with successful out-of-sample generalization is not necessarily statistically significant. The distributional properties of an effect important for high statistical significance and for successful generalization are not identical (Lo et al., 2015). Additionally, classical inference is a judgment about an entire data sample, whereas predictive inference can be applied to single datapoints. The goal and permissible conclusions of a formal inference therefore are conditioned by the adopted statistical framework (Feyerabend, 1975). This is routinely exploited in drug development cycles with predictive inference for the early discovery phase and classical inference for the clinical trials phase. A similar back and forth between applying inductive learning algorithms and deductive hypothesis testing of the discovered candidate structure could and should also become routine in imaging neuroscience. Awareness of the discussed cultural gap is important to keep pace with the increasing information granularity of acquired neuroimaging repositories. Ultimately, statistical inference is a heterogenous concept.

**Acknowledgments**

The present paper did not result from isolated contemplations by a single person. Rather, it emerged from several thought milieus with different thought styles and opinion systems. The author cordially thanks the following people for valuable discussion and precious contributions to the present paper (in alphabetical order): Olavo Amaral, Patrícia Bado, Jérémy Lefort-Besnard, Kay Brodersen, Elvis Dohmatob, Guillaume Dumas, Michael Eickenberg, Simon Eickhoff, Denis Engemann, Can Ergen, Alexandre Gramfort, Olivier Grisel, Carl Hacker, Michael Hanke, Lukas Hensel, Thilo Kellermann, Jean-Rémi King, Robert Langner, Daniel Margulies, Jorge Moll, Zeinab Mousavi, Carolin Mößnang, Rico Pohling, Andrew Reid, João Sato, Bertrand Thirion, Gaël Varoquaux, Marcel van Gerven, Virginie van Wassenhove, Klaus Willmes, Karl Zilles.